\documentclass[twoside,11pt]{article}

\usepackage{blindtext}
\usepackage[abbrvbib,preprint]{jmlr2e}

\usepackage[utf8]{inputenc} 
\usepackage[T1]{fontenc}    
\usepackage{hyperref}       
\usepackage{url}            
\usepackage{booktabs}       
\usepackage{amsfonts}       
\usepackage{nicefrac}       
\usepackage{microtype}      
\usepackage{xcolor}         
\usepackage{amssymb}
\usepackage{amsmath}
\usepackage[bb=dsserif]{mathalpha}
\usepackage{bm}
\usepackage{caption}
\usepackage{footnote}
\usepackage{lmodern}
\usepackage{mathtools}
\usepackage{multirow}
\usepackage{subcaption}
\usepackage{lastpage}
\usepackage{algorithm}
\usepackage{algpseudocode}
\usepackage{amsmath}

\DeclareMathOperator*{\argmin}{arg\,min}

\firstpageno{1}

\begin{document}

\title{Series-wise Cross-validation for Global Time Series Forecasting}

\author{\name Vitor Cerqueira 
        \email vitorc@dei.uc.pt \\
       \addr University of Coimbra, CISUC/LASI, DEI\\
       Coimbra, Portugal
       \AND
       \name Mariana Oliveira 
       \email mariana.oliveira@fe.up.pt \\
       \addr Faculty of Engineering, University of Porto\\ Porto, Portugal
       \AND
       \name Luis Roque
       \email luis\_roque@live.com \\
       \addr Faculty of Engineering, University of Porto\\ Porto, Portugal
       }

\editor{My editor}

\maketitle

\begin{abstract}

We consider the problem of cross-validation (CV) in time series forecasting, with a focus on global forecasting models based on deep learning.
The prevailing CV strategy in forecasting is to partition data across time with the last part of each series used for testing. With global models trained on collections of time series, however, a fundamentally different approach becomes possible: partitioning across \emph{series} and time. We formalize \emph{series-wise} CV for global forecasting models and benchmark several instantiations, including $K$-fold CV, bootstrapping, and Monte Carlo CV, on 11 datasets with 10 neural network architectures. We evaluate CV procedures along five dimensions: performance estimation, overoptimism rate, model selection accuracy, regret, and downstream forecasting performance. We withhold a random fraction of series from training so that they appear only in the test set. This protocol simulates the emergence of new series not seen during training, a common scenario in global forecasting. Under this protocol, our results reveal a nuanced trade-off: series-wise CV yields better downstream forecasting accuracy by guiding model selection toward configurations that generalize across both time and series, while time-wise CV retains a slight advantage for estimating out-of-sample error. 

\end{abstract}


\section{Introduction}\label{sec:intro}

\subsection{Context and Motivation}

Cross-validation (CV) is the predominant approach for estimating performance and selecting models in machine learning~\cite{arlot2010survey,bates2024cross}. Evaluating models in held-out data yields more reliable estimates than in-sample or goodness-of-fit measures~\cite{tashman2000out}, and enables the selection of algorithms that generalize well in unseen data.

\begin{figure}[!t]%
    \centering
    \subfloat[\centering Time-wise CV]{{\includegraphics[width=.37\textwidth]{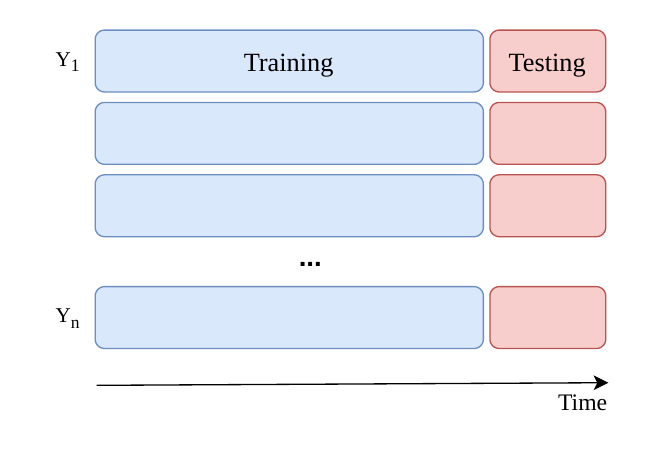} }}%
    \quad
    \subfloat[\centering Series-wise CV]{{\includegraphics[width=.58\textwidth]{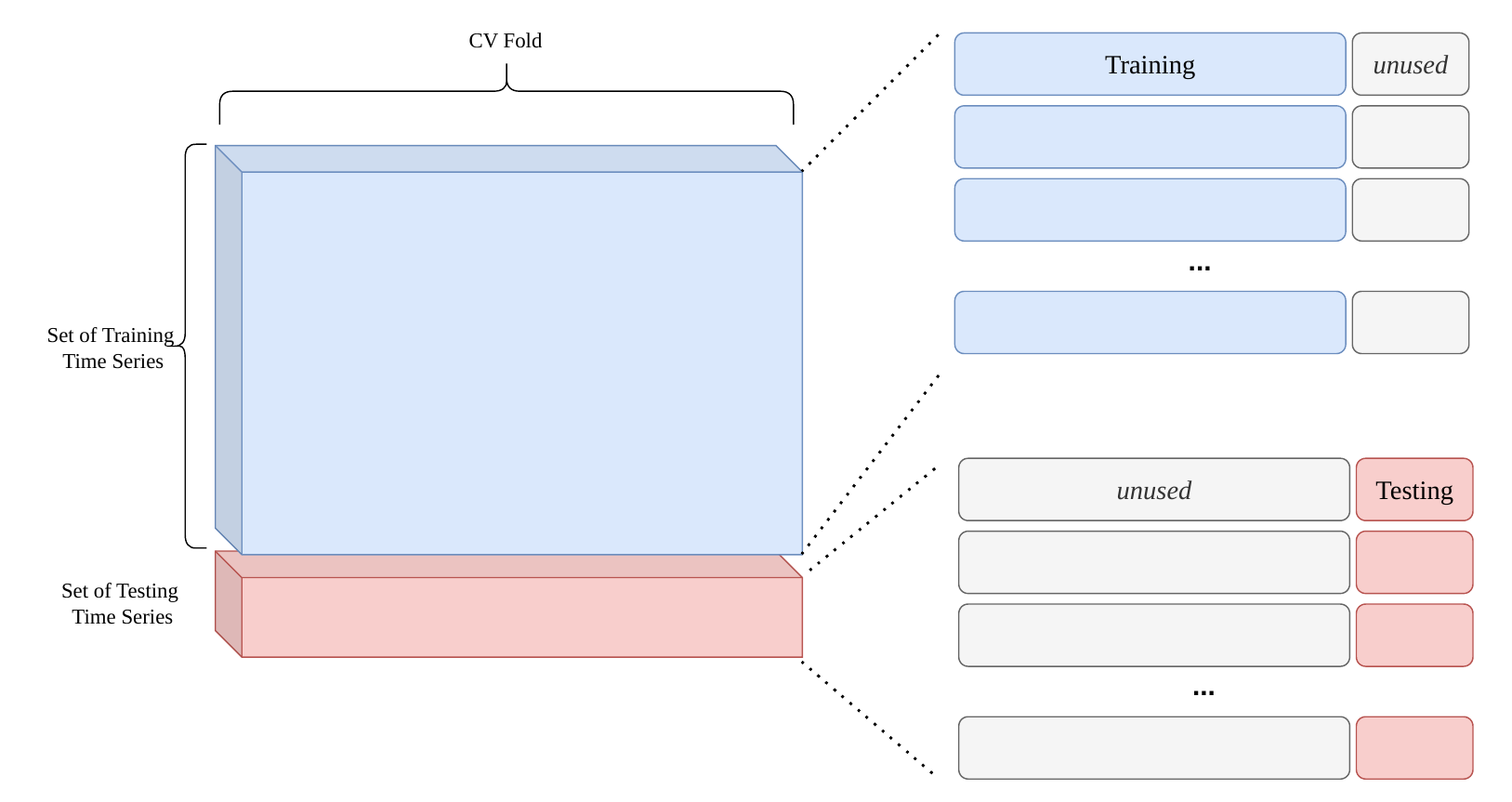}}}%
    \caption{Illustration of time-wise CV (a) and series-wise CV (b). In time-wise CV, all series contribute to both training and testing, with only the temporal dimension used for splitting. In series-wise CV, the split is across \emph{series} and time: one subset is used for training, a disjoint subset for testing, and both are further split in time to preserve temporal order.}%
    \label{fig:diagrams}%
\end{figure}

With the rise of deep learning in forecasting~\cite{aksu2024gift}, state-of-the-art models are increasingly \emph{global}: a single model is trained on a collection of time series, rather than fitting one model per series as in classical methods such as ARIMA. 
In both global and local approaches, the standard evaluation protocol is to hold out the last observations of a given series for testing. We call this \emph{time-wise CV} (Figure~\ref{fig:diagrams}a).  

Time-wise CV preserves temporal order and emulates a realistic deployment scenario, but training and validation data come from the same series and differ only in time. Thus, raising questions about generalization to entirely new series, as opposed to temporal extrapolation on series whose history was used for training.
Alternative splitting strategies, such as blocking~\cite{bergmeir2012use} or embargoing~\cite{racine2000consistent}, have been explored for single time series; however, time-wise splitting yields better performance estimates in that setting~\cite{cerqueira2020evaluating}. Moreover, most previous work about time series forecasting has focused on local models fit to individual series, leaving a gap in our understanding of CV for global forecasters trained on many series.

Datasets comprising multiple time series open a new possibility: partitioning across \emph{series} as well as time. In \emph{series-wise} CV (Figure~\ref{fig:diagrams}b), models are trained on the history of one subset of series and evaluated on the future of a disjoint subset. This dual partitioning, across both the series and temporal dimensions, is the focus of this paper.

\subsection{Objectives and Contributions}

We hypothesize that series-wise CV can improve downstream forecasting accuracy by encouraging the selection of models that generalize across both time and the space of time series. In time-wise CV, validation data come from the same series used for training, differing only in time; the selected model is therefore optimized for temporal extrapolation on known dynamics. Series-wise CV, by contrast, evaluates candidates on entirely unseen series, rewarding configurations that not only generalize well across time but also transfer well to new patterns. 
This objective is especially relevant as time series forecasting has moved from purely local models to global models and, more recently, to time series foundation models~\cite{ansari2024chronos}. These models are applied to populations of series that evolve over time, in which new entities arrive and the dynamics of existing series shift~\cite{kim2021reversible}.

This paper makes the following contributions.

\begin{enumerate}
    \item We formalize series-wise CV for global forecasting models and present several instantiations, such as $K$-fold CV or bootstrapping, that partition datasets across both series and time.

    \item We design a benchmark that evaluates CV procedures along five dimensions: (i)~\emph{performance estimation} (how close CV estimates are to the true out-of-sample error); (ii) \emph{overoptimism rate} (how often the actual loss is greater than the estimated one); (iii)~\emph{selection accuracy} (how often CV picks the best algorithm); (iv)~\emph{regret} (gap between the selected model and the best attainable); and (v)~\emph{downstream forecasting performance}, treating each CV procedure as part of the full forecasting pipeline.

    \item We experiment on 11 benchmark time series datasets using 10 well-established neural network architectures. Our evaluation protocol simulates the emergence of new series in the test set that were not seen during training, a common scenario in global forecasting.
    The results reveal a nuanced trade-off: series-wise CV yields better downstream forecasting performance while time-wise CV retains a slight edge for performance estimation. In other words, series-wise CV guides model selection toward configurations that generalize better, even though time-wise splits produce CV estimates closer to out-of-sample error.
\end{enumerate}

\noindent The code to reproduce the experiments is available at \url{https://github.com/vcerqueira/experiments-tscv}.

\section{Background}

This section reviews global time series forecasting and the standard practice of CV for temporally ordered data. In forecasting, CV holds out the future of a series so that temporal order is preserved. Studies of this approach have focused on individual series, so how to evaluate models trained jointly on many series remains an open question, especially in scenarios where new series emerge in the dataset.

\subsection{Time series forecasting}

Let \( \mathcal{Y} = \{ Y^{i} \}_{i=1}^n \) denote a collection of $n$ time series. Each series \( Y^{i} = \{ y^{i}_j \}_{j=1}^{n_i} \) is a sequence of numeric observations indexed by time $j$, with $n_i$ the length of the $i$-th series. The forecasting task is to predict the next $h$ values of a series given its past; we write the target as $Y^{i}_{t+1:t+h} = \{ y^{i}_{t+1}, \dots, y^{i}_{t+h} \}$.

Global forecasting models train a single model on $\mathcal{Y}$, exploiting shared structure across series and often improving accuracy and data efficiency relative to local (per-series) models~\cite{januschowski2020criteria, godahewa2021ensembles}. Using a context window of length $p$, each series is turned into input-output pairs via time-delay embedding~\cite{bontempi2013machine}: at each time $t$, the input is the window $X^{i}_t \in \mathbb{R}^p$ and the output is the subsequent horizon $Y^{i}_{t+1:t+h} \in \mathbb{R}^h$:
\begin{equation}
X^{i}_t = \{ y^{i}_{t-p+1}, \dots, y^{i}_t \}, \qquad 
Y^{i}_{t+1:t+h} = \{ y^{i}_{t+1}, \dots, y^{i}_{t+h} \}.
\end{equation}
Training amounts to fitting a global function $f_\theta : \mathbb{R}^p \rightarrow \mathbb{R}^h$ so that $\hat{Y}^{i}_{t+1:t+h} = f_\theta(X^{i}_t)$ approximates the true future values in all series and time steps.

Deep learning offers a wide variety of approaches to time series forecasting, including recurrent, convolutional, attention-based, and feed-forward designs. In large-scale benchmark competitions and standardized accuracy studies, global models based on deep learning have repeatedly achieved state-of-the-art results relative to classical statistical methods~\cite{aksu2024gift,makridakis2022m5}, which has driven their adoption for forecasting.

\subsection{Time series cross-validation}

The key design choice in CV is how to partition the data into training and validation sets.
For i.i.d.\ data, CV is straightforward: observations can be randomly assigned to training and validation folds~\cite{arlot2010survey}. Time series violate the i.i.d. assumption, so the partitioning of observations must take temporal order into account. Many variants of randomized CV have been proposed to recover the data efficiency of $k$-fold CV (e.g., blocking~\cite{bergmeir2012use} or embargoing~\cite{racine2000consistent}), but in most forecasting settings it is widely held that credible evaluation requires preserving temporal order, with test observations corresponding to more recent observations than those in the training set~\cite{cerqueira2020evaluating,hewamalage2023forecast}.

Several studies have benchmarked CV strategies under temporal constraints for forecasting and related dependent-data settings~\cite{bergmeir2012use,bergmeir2018note,cerqueira2020evaluating,oliveira2021evaluation}. Much of it focuses on whether CV recovers the true generalization error; a smaller body also studies how the choice of CV affects model selection~\cite{cerqueira2023model}. All of this work on univariate time series forecasting considers \emph{single} time series (local models fit per series)~\cite{cerqueira2020evaluating} or assumes spatial dependence between series~\cite{oliveira2021evaluation}. 
To our knowledge, there are no existing studies of global forecasters trained jointly in many series in the context addressed in this paper, which motivates the series-wise perspective in the next section.

\section{Series-wise Time Series CV}\label{sec:swcv}

This section formally introduces series-wise time series CV, describing its key principles and nested application. We also present how state-of-the-art estimators, such as bootstrapping or Monte Carlo simulation, can be applied for series-wise CV. 

\subsection{Definition and framework}\label{sec:swcv_preliminaries}

The fundamental principle behind series-wise CV is that, at any given fold, the model is trained on one subset of time series and evaluated on another, disjoint subset, while also respecting temporal order. 
This separation provides a stricter evaluation than time-wise CV, where the model has already observed each series during training. 
Thus series-wise CV evaluates generalization not only across time but also across the space of time series, a desirable property when the goal is to select models that perform well on unseen series or under distribution shift.

Formally, series-wise CV is grounded on partitioning the collection \( \mathcal{Y} \) into disjoint subsets. Let the dataset \( \mathcal{Y} = \{ Y^{i} \}_{i=1}^n \) be partitioned into non-overlapping subsets of series indices:
\[
\mathcal{I}_{\text{train}} \subset \{1, \dots, n\}, \quad \mathcal{I}_{\text{test}} = \{1, \dots, n\} \setminus \mathcal{I}_{\text{train}},
\]
such that the training and test sets are defined as:
\[
\mathcal{Y}_{\text{train}} = \{ Y^{i} : i \in \mathcal{I}_{\text{train}} \}, \quad \mathcal{Y}_{\text{test}} = \{ Y^{i} : i \in \mathcal{I}_{\text{test}} \}.
\]
Crucially, $\mathcal{Y}_{\text{train}} \cap \mathcal{Y}_{\text{test}} = \emptyset$, ensuring no time series appears in both subsets.

In addition to partitioning across series, series-wise CV preserves temporal constraints by further splitting each time series across time. For each series $Y^{i}$ with length $n_i$, we define a temporal split point $\tau_i < n_i$ such that:
\[
Y^{i}_{\text{history}} = \{ y^{i}_1, \dots, y^{i}_{\tau_i} \}, \quad Y^{i}_{\text{future}} = \{ y^{i}_{\tau_i+1}, \dots, y^{i}_{n_i} \}.
\]
During training, the model $f_\theta$ is fitted using input-output pairs constructed from $\{ Y^{i}_{\text{history}} : i \in \mathcal{I}_{\text{train}} \}$. During evaluation, the model is tested on forecasting $Y^{i}_{\text{future}}$ for series $i \in \mathcal{I}_{\text{test}}$, using only the corresponding $Y^{i}_{\text{history}}$ as context.
This dual partitioning, across both series and time, distinguishes series-wise CV from conventional time-wise CV, where all series contribute to both training and testing, with only the temporal dimension used for splitting. 
The series-wise CV framework can be expressed as:
\begin{equation}
\hat{Y}^{i}_{\text{future}} = f_\theta(Y^{i}_{\text{history}}), \quad \forall \, i \in \mathcal{I}_{\text{test}},
\end{equation}
where $f_\theta$ is trained exclusively on series from $\mathcal{I}_{\text{train}}$.

The performance estimate for a given fold is computed by aggregating forecasting errors over the test series:
\begin{equation}
\hat{\mathcal{L}} = \frac{1}{|\mathcal{I}_{\text{test}}|} \sum_{i \in \mathcal{I}_{\text{test}}} \ell\left( Y^{i}_{\text{future}}, \hat{Y}^{i}_{\text{future}} \right),
\end{equation}
where $\ell(\cdot, \cdot)$ denotes a loss function. When a strategy produces multiple folds, the overall performance estimate is typically the mean of the per-fold losses $\hat{\mathcal{L}}$ over all folds.

\subsection{Strategies}\label{sec:swcv_listofstrats}

The specific manner in which $\mathcal{I}_{\text{train}}$ and $\mathcal{I}_{\text{test}}$ are constructed defines different series-wise CV strategies, such as the following:

\begin{itemize}
    \setlength{\itemsep}{0pt}
    \setlength{\parsep}{0pt}
    \setlength{\parskip}{0pt}
    
    \item \texttt{Holdout}: The classical and simplest approach is to partition the $n$ series into a training set and a test set by assigning a random subset of series (e.g., a fixed percentage, often 60--80\%) to $\mathcal{I}_{\text{train}}$ and the remainder to $\mathcal{I}_{\text{test}}$. Holdout is most reliable when the number of series is large enough that the held-out test set is representative of the population~\cite{cerqueira2020evaluating}.

    \item Repeated Holdout (\texttt{K-Holdout}): A single holdout split can yield unrepresentative train or test sets. Repeating the holdout $K$ times with different random partitions and averaging the performance estimates yields more stable and reliable estimates.

    \item Monte Carlo CV (\texttt{K-MC-CV}): For each of $K$ repetitions, a random subset of series is drawn for training, a disjoint random subset for testing, and the remaining series are left unused in that repetition. This differs from \texttt{K-Holdout}, where the training and test sets together cover all $n$ series in every repetition. Different variants of Monte Carlo CV have shown competitive performance estimates in other data-dependent scenarios~\cite{arlot2010survey}.

    \item \texttt{Bootstrap}: Whereas holdout uses sampling without replacement, bootstrap constructs $\mathcal{I}_{\text{train}}$ by sampling $n$ series \emph{with} replacement from $\{1,\ldots,n\}$; $\mathcal{I}_{\text{test}}$ is the set of indices that were not selected (out-of-bag) (on average, 63.2\% of series appear at least once in the training set~\cite{tibshirani1993introduction}).

    \item \texttt{K-Bootstrap}: As with Holdout, repeating the bootstrap procedure $K$ times and averaging reduces variance in the performance estimate.

    \item $K$-fold Cross-validation (\texttt{Kf-CV}): Holdout and bootstrap do not guarantee that every series is used for testing. In $K$-fold CV, the series indices $\{1,\ldots,n\}$ are partitioned into $K$ disjoint blocks. Each fold uses one block as $\mathcal{I}_{\text{test}}$ and the union of the remaining blocks as $\mathcal{I}_{\text{train}}$, so every series is tested exactly once and all data contribute to the estimate. 
\end{itemize}

\begin{figure}[!t]%
    \centering
    \subfloat[\centering K-fold CV with k=5]{{\includegraphics[width=.95\textwidth]{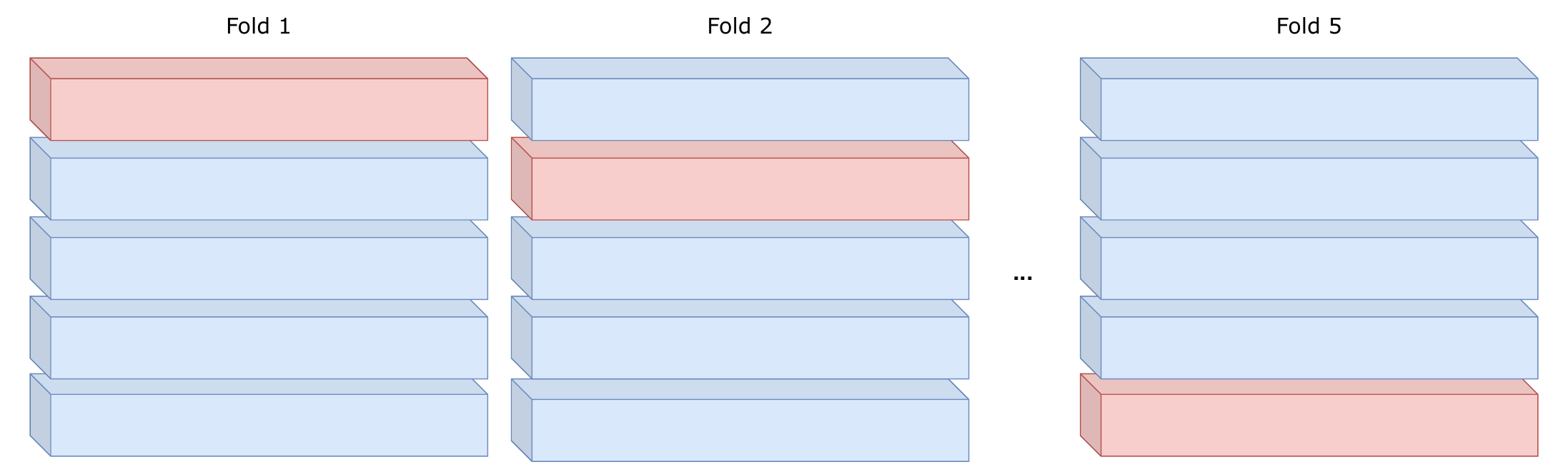}}}%
    
    \subfloat[\centering Holdout]{{\includegraphics[width=.29\textwidth]{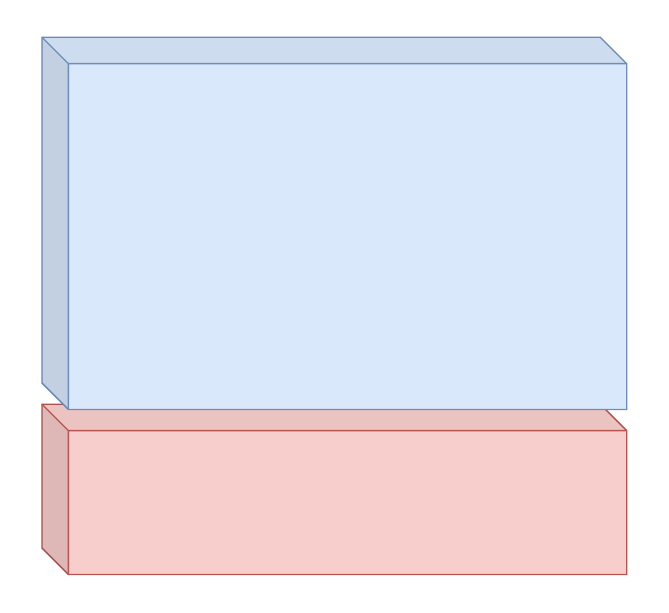} }}%
    \quad
    \subfloat[\centering Monte Carlo CV (example fold)]{{\includegraphics[width=.29\textwidth]{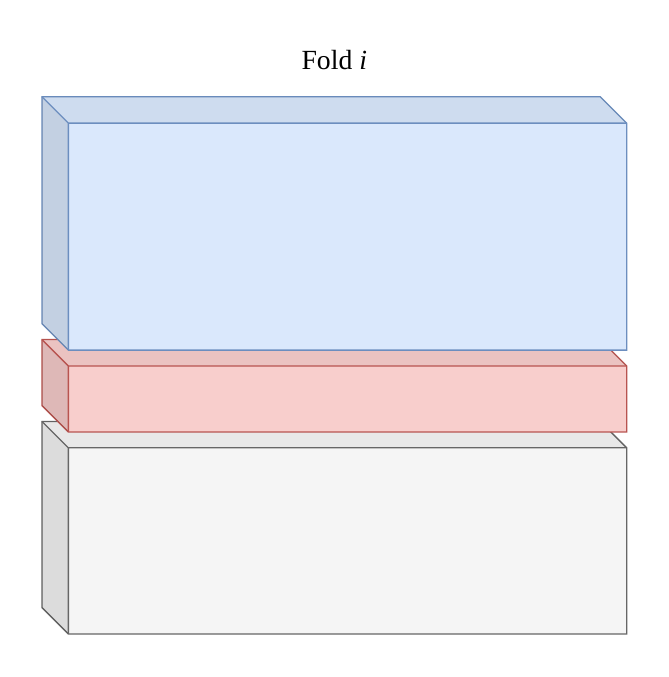} }}%
    \quad
    \subfloat[\centering Bootstrap]{{\includegraphics[width=.29\textwidth]{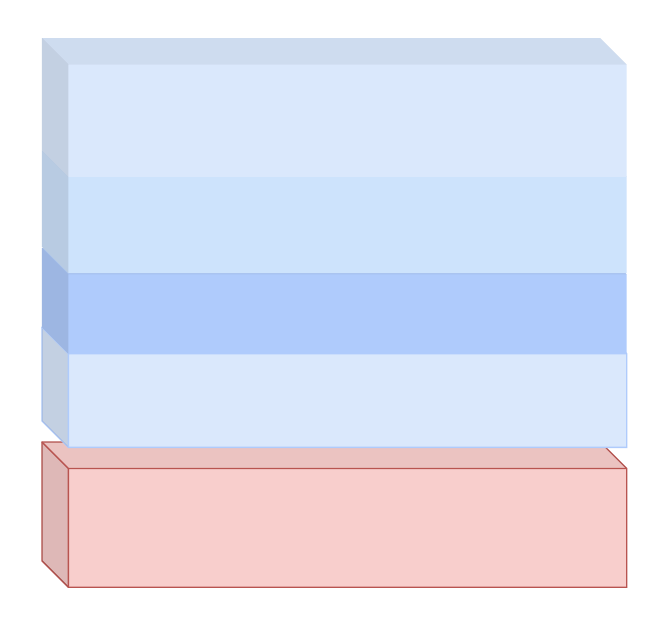}}}%
    \caption{Illustration of series-wise CV partitioning schemes. Each horizontal bar represents a time series; blue indicates training data and red indicates test data. (a) $K$-fold CV with $K{=}5$: the series are split into 5 disjoint folds, and each fold serves as the test set exactly once. (b) Holdout: a single random split assigns most series to training and the remainder to testing. (c) Monte Carlo CV: a random subset of series is used for training and a disjoint subset for testing; gray indicates unused series in that iteration. (d) Bootstrap: training series are sampled with replacement (darker shades indicate that a series may appear multiple times in the training set) and out-of-bag series form the test set.}%
    \label{fig:series_wise_cvs}%
\end{figure}

Figure~\ref{fig:series_wise_cvs} illustrates these partitioning schemes; blue regions denote training data and red regions denote test data.

\subsection{Nested series-wise CV}\label{sec:swcv_nested}

\begin{figure}[bt]
    \centering
    \includegraphics[width=\textwidth, trim=0cm 0cm 0cm 0cm, clip=TRUE]{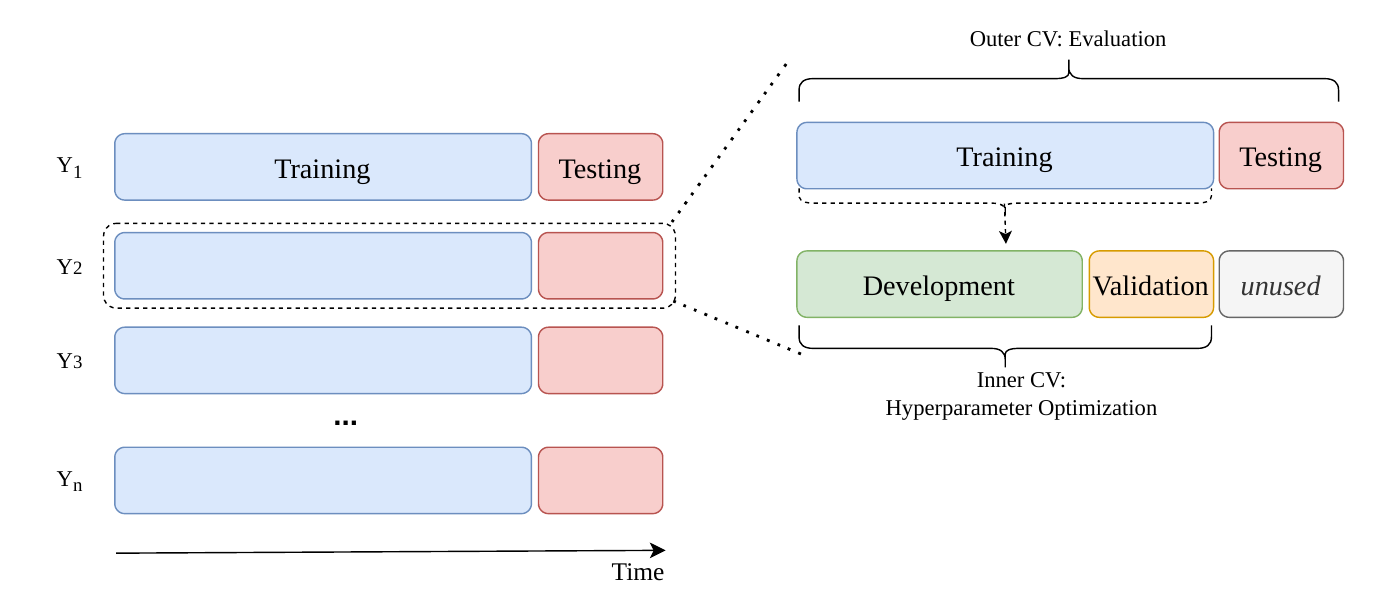}
    \caption{Nested time-wise CV. The outer loop partitions data across time for performance estimation; the inner loop further splits the outer training portion across time to select hyperparameters. All series contribute to both training and testing at every level.}
    \label{fig:typical_approach}
\end{figure}

In practice, CV serves both performance estimation and model selection: we use CV to estimate how well a model generalizes and to choose among algorithms or hyperparameter configurations~\cite{arlot2010survey}. Nested CV separates these roles by using an \emph{outer} loop for performance estimation and an \emph{inner} loop for model or hyperparameter selection on each outer training set. Figure~\ref{fig:typical_approach} shows the time-wise approach for nested CV.

In nested series-wise CV, the outer loop applies one of the strategies from Section~\ref{sec:swcv_listofstrats} to partition the collection $\mathcal{Y}$ into series-wise folds. For each outer fold, $\mathcal{I}_{\text{train}}$ and $\mathcal{I}_{\text{test}}$ are defined as before; the model is fitted on the training series and evaluated on the test series. Within each outer fold, however, we must also select a model or tune hyperparameters using only the data from the outer training series. 

\begin{figure}[tb]
    \centering
    \includegraphics[width=\textwidth, trim=0cm 0cm 0cm 0cm, clip=TRUE]{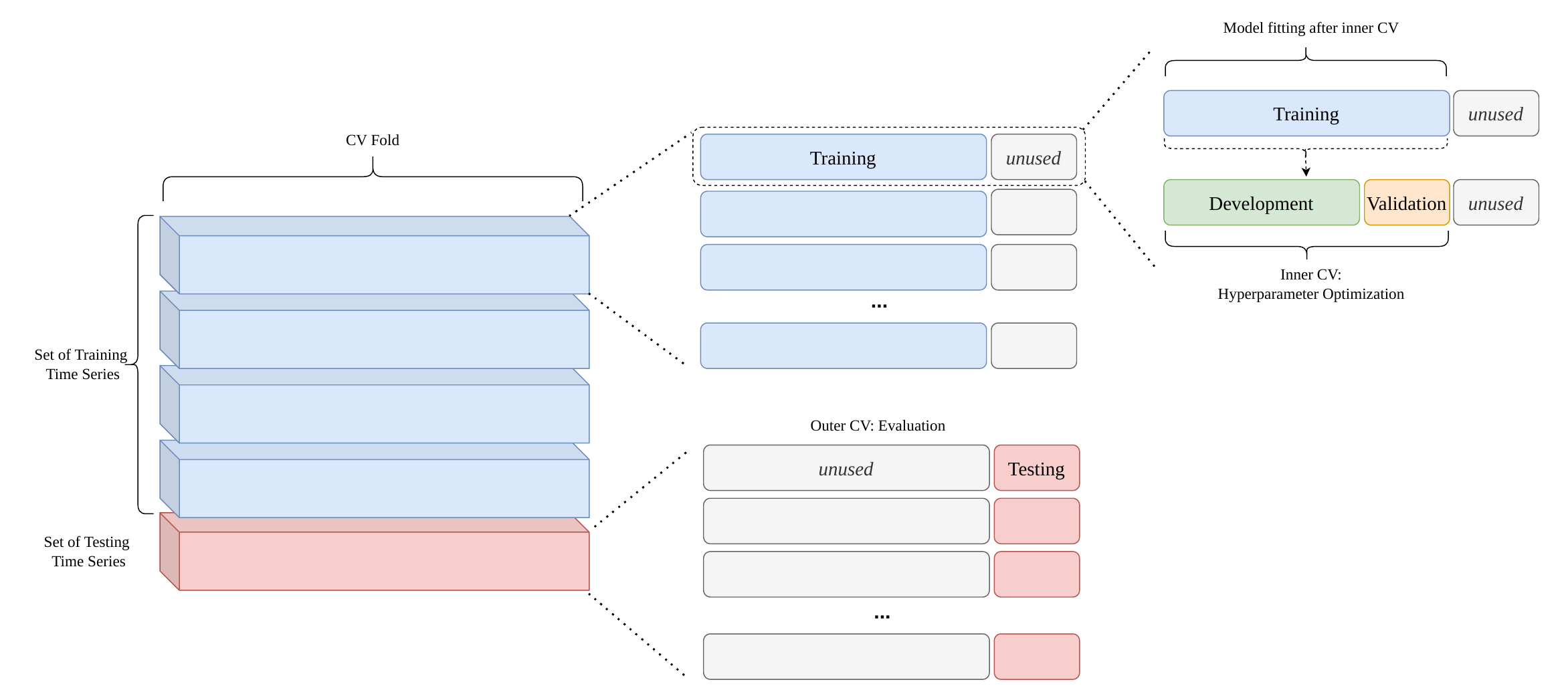}
    \caption{Nested series-wise CV. The outer loop partitions data across series and time for performance estimation: training series and test series are disjoint. The inner loop applies a time-wise holdout on the outer training series to select hyperparameters. This ensures that outer-test series are never seen during training or tuning.}
    \label{fig:series_wise_cv_approach}
\end{figure}

The inner loop can partition the outer training series across time only, or across both series and time. 
In this work we adopt the simpler option: time-wise holdout within the inner loop. Concretely, for each outer fold the most recent observations of $\mathcal{I}_{\text{train}}$ serve as the inner validation set, and the remaining history serves as the inner training set. Once the best hyperparameters are selected, the model is retrained on all of $\mathcal{I}_{\text{train}}$ and evaluated on $\mathcal{I}_{\text{test}}$ to produce one term of the outer performance estimate. This design mirrors standard nested CV practice, where the inner procedure is often a single holdout even when the outer loop is $K$-fold CV, effectively balancing statistical rigor with computational tractability. For the comparison of CV approaches in Section~\ref{sec:experiments}, this choice also isolates the effect of the outer CV strategy and enables a fair comparison across outer-loop partitioning methods.
Figure~\ref{fig:series_wise_cv_approach} illustrates the complete workflow. The outer series-wise folds guarantee that performance is estimated on series unseen during both training and tuning.

\section{Experiments}\label{sec:experiments}

This section describes the experimental setup designed to benchmark the performance of different series-wise time series CV approaches for performance estimation and model selection of global forecasting models. The evaluation pipeline is described in Section~\ref{sec:performance_estimation_evaluation}. The datasets and learning algorithms are described in Sections~\ref{sec:datasets} and~\ref{sec:learning_algorithms}, respectively.

\subsection{Performance estimation evaluation methodology}\label{sec:performance_estimation_evaluation}

\paragraph{Test set for CV methods}

To benchmark the performance of different CV approaches, we design an evaluation procedure that emulates a realistic scenario by holding out a test set of observations of each time series. The dataset $\mathcal{Y}$ is partitioned across time into two disjoint subsets, $\mathcal{Y}_{in-set}$ and $\mathcal{Y}_{out-set}$. The out-set $\mathcal{Y}_{out-set}$ is constructed by withholding the last $2 \times h$ observations from each time series, while the in-set $\mathcal{Y}_{in-set}$ consists of all earlier observations:
\begin{equation}\label{eq:temporal-split}
    \mathcal{Y}_{in-set} = \left\{ y^{(i)}_{1: n_i - 2h} \right\}_{i=1}^n,
    \qquad
    \mathcal{Y}_{out-set} = \left\{ y^{(i)}_{n_i - 2h + 1: \, n_i} \right\}_{i=1}^n.
\end{equation}

For all CV procedures applied to $\mathcal{Y}_{in-set}$, the temporal test portion in each fold spans $\tau = h$ observations, matching the forecasting horizon.
The in-set is used for training, hyperparameter tuning, and performance estimation by CV. The out-set is held out entirely and only used for the final model evaluation to compute the true forecasting loss. The $2 \times h$ window is consumed in a rolling origin manner~\cite{hewamalage2023forecast}, predicting the next $h$ observations at each time step. In other words, the out-set is the test set for CV procedures, enabling the assessment of the performance estimates, selection accuracy, and forecasting performance.

Equation~\eqref{eq:temporal-split} splits the data across time only, so every series contributes to both the in-set and the out-set. To emulate a population in which new series appear only at test time, we additionally restrict the in-set to a random subset of series. Let $\mathcal{S} \subseteq \{1, \dots, n\}$ be a random subset of series indices with $|\mathcal{S}| = \lfloor \rho n \rfloor$ and $\rho = 0.7$. The in-set used by the CV procedures retains the earlier observations of these series only, while the out-set is unchanged:
\begin{equation}\label{eq:series-split}
    \mathcal{Y}_{in-set} = \left\{ y^{(i)}_{1: n_i - 2h} \right\}_{i \in \mathcal{S}},
    \qquad
    \mathcal{Y}_{out-set} = \left\{ y^{(i)}_{n_i - 2h + 1: \, n_i} \right\}_{i=1}^n.
\end{equation}
The remaining series, $\{1, \dots, n\} \setminus \mathcal{S}$, appear only in the out-set and are therefore unseen during training, tuning, and model selection, emerging as new series at test time. This emulates a setting where the population of series evolves over time, precisely where series-wise CV might shine and increasingly common in the age of global forecasting models and time series foundation models~\cite{ansari2024chronos}. Setting $\rho = 1$ recovers the split in Equation~\eqref{eq:temporal-split}, where all series appear in the in-set; we also report results for this case for completeness.

\paragraph{Nested CV}

All procedures use nested CV on the $\mathcal{Y}_{in\text{-}set}$ only. For each candidate forecaster, the \emph{inner} loop applies a time-wise split (cf. Section~\ref{sec:swcv_nested}) to choose hyperparameters. The \emph{outer} loop follows the chosen series-wise (or time-wise) CV scheme: for each outer fold, we retrain the model with the inner-selected hyperparameters on all outer-training data, evaluate on the outer-test portion of the in-set, and record the loss. Averaging these outer-fold losses yields the performance estimate for that forecaster under the given CV procedure. To score the CV procedure itself, we retrain each forecaster using its best hyperparameters on the \emph{entire} in-set, according to each estimator, and evaluate on $\mathcal{Y}_{out\text{-}set}$.

\paragraph{Forecasting performance metric}

We use MASE as the forecasting performance metric~\cite{makridakis2018m4}, with a global scaling variant. Standard MASE scales a model’s MAE by a seasonal naive MAE; here, we define the scaling factor ($Q$) as the MAE of the seasonal naive baseline computed once over all series available in the in-set. For each CV fold, the numerator is the model MAE on that fold’s evaluated series/horizons. Using a fixed global ($Q$) keeps scores comparable across folds and CV procedures, even when evaluated series differ.

\paragraph{CV evaluation metrics}

We formalize the evaluation of CV procedures using the framework of the Combined Algorithm Selection and Hyperparameter optimization (CASH) problem~\cite{thornton2013auto}. Given a set of learning algorithms $\mathcal{A}$ and a dataset $\mathcal{Y}$, the goal is to select the algorithm $A^\star \in \mathcal{A}$ that minimizes the expected loss on unseen data. 
For each algorithm $A \in \mathcal{A}$, a CV procedure produces an estimated MASE loss $\hat{\mathcal{L}}(A)$ computed on the in-set $\mathcal{Y}_{in\text{-}set}$. The true loss $\mathcal{L}(A)$ is the MASE error incurred by algorithm $A$ when evaluated on the held-out out-set $\mathcal{Y}_{out\text{-}set}$.
The CV procedure selects the algorithm with the lowest estimated loss, $A_{sel} = \argmin_{A \in \mathcal{A}} \hat{\mathcal{L}}(A)$, while the oracle selects the algorithm with the lowest true loss, $A^\star = \argmin_{A \in \mathcal{A}} \mathcal{L}(A)$.
Based on these quantities, we define the following evaluation metrics across a collection of $D$ datasets:

\begin{itemize}
    \item \textbf{MAPEE} (Mean Absolute Performance Estimation Error) (e.g.~\cite{cerqueira2020evaluating}): The average absolute difference between the estimated MASE loss and the true MASE loss, averaged across all algorithms and datasets:

    \[
    \text{MAPEE} = \frac{1}{D}\frac{1}{|\mathcal{A}|} \sum_{d=1}^{D}  \sum_{A \in \mathcal{A}} \left| \hat{\mathcal{L}}_d(A) - \mathcal{L}_d(A) \right|.
    \]

    \item \textbf{Overoptimism Rate}: The proportion of dataset and model pairs in which the true MASE loss was under-estimated by the CV procedure, considering that in certain applications under-estimating the error may be more costly than over-estimating it:

    \[
    \text{Overoptimism} = \frac{1}{D}\frac{1}{|\mathcal{A}|} \sum_{d=1}^{D}  \sum_{A \in \mathcal{A}} \mathbb{1} \left[ \hat{\mathcal{L}}_d(A) < \mathcal{L}_d(A) \right].
    \]

    \item \textbf{Selection Accuracy}: The proportion of datasets where the CV-selected model is also the oracle-selected model:
    \[
    \text{Accuracy} = \frac{1}{D} \sum_{d=1}^{D} \mathbb{1}\left[ A_{sel}^{(d)} = A^{\star(d)} \right].
    \]

    \item \textbf{Regret}: The average difference between the true loss of the selected model and the true loss of the oracle model. Regret captures robustness: a procedure may occasionally miss the optimum, yet still select an accurate alternative~\cite{cerqueira2023model}.\footnote{Because each CV approach tunes hyperparameters independently, the ``best model'' can differ across procedures; we therefore compute regret relative to each procedure's own candidate set.}
    \[
    \text{Regret} = \frac{1}{D} \sum_{d=1}^{D} \left( \mathcal{L}_d(A_{sel}^{(d)}) - \mathcal{L}_d(A^{\star(d)}) \right).
    \]

    \item \textbf{Downstream Forecasting Performance}: Ultimately, a CV procedure is part of a forecasting workflow. For a given CV procedure $g$, we apply it to the in-set to select an algorithm $A_{sel}^{g}$, retrain on the entire in-set, and evaluate on the out-set according to MASE. This metric directly captures the end-to-end value of a CV procedure: a method that consistently selects well-generalizing models will achieve better downstream forecasting accuracy.
\end{itemize}

We repeat the complete evaluation procedure for 3 different seeds (123, 456, 789) to assess the robustness of the results for distinct initializations. We report the average score in the main body of the paper, while the appendix contains the per-seed results.

\begin{figure}[bt]
    \centering
    \includegraphics[width=\textwidth, trim=0cm 0cm 0cm 0cm, clip=TRUE]{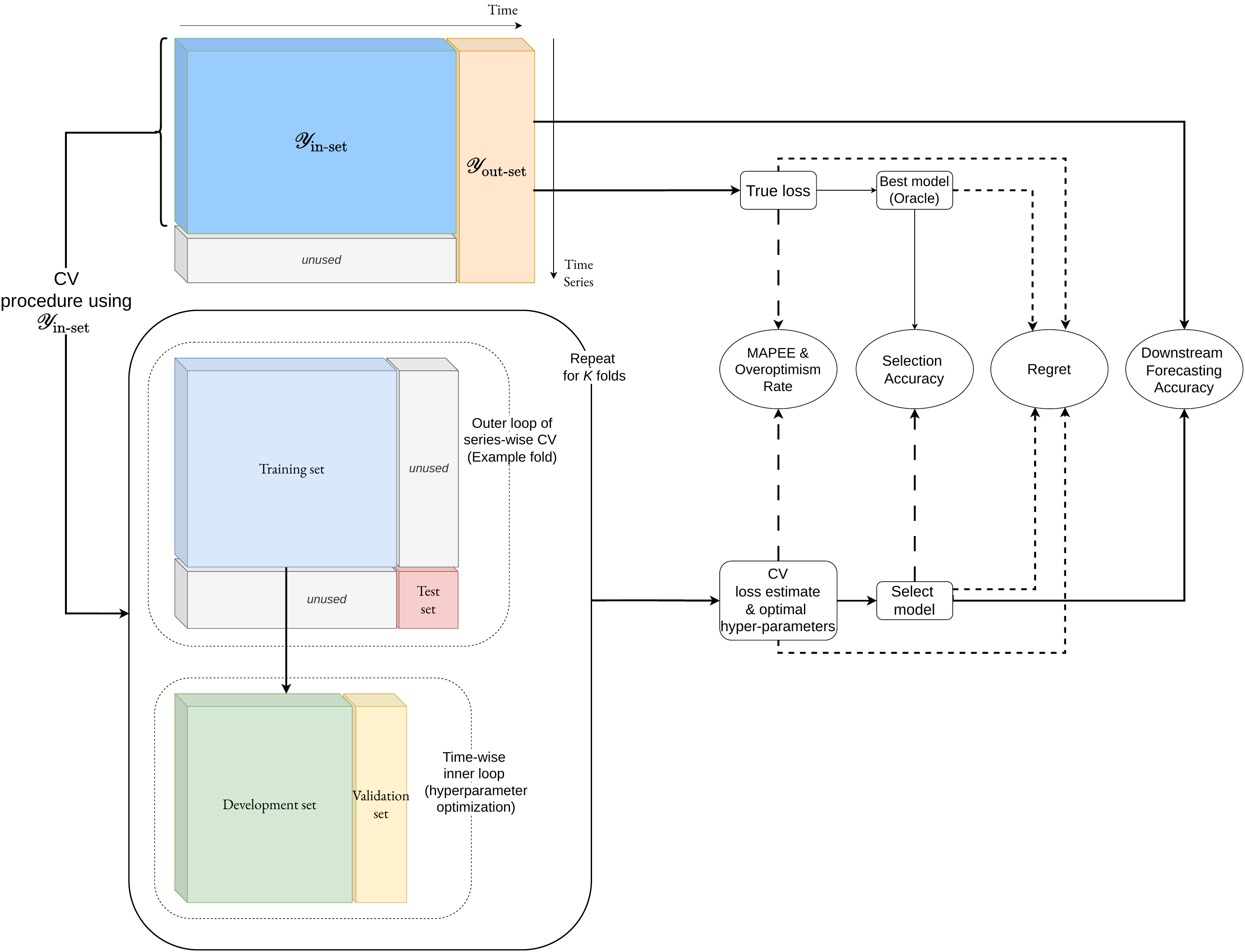}
    \caption{Evaluation protocol and flow of information for scoring a CV procedure.}
    \label{fig:diagram_protocol}
\end{figure}

The overall protocol for scoring a CV procedure is illustrated in Figure~\ref{fig:diagram_protocol}, detailing the flow of information required for computing each metric. In the outer fold, the plot shows an example partitioning. Different strategies can be applied based on the list in Section~\ref{sec:swcv_listofstrats}.

\subsection{Datasets}\label{sec:datasets}

We conduct experiments on 11 benchmark datasets spanning diverse domains, sampling frequencies, and forecasting horizons: Tourism~\cite{athanasopoulos2011tourism} (monthly and quarterly), M1~\cite{makridakis1982accuracy} (monthly and quarterly), M3~\cite{makridakis2000m3} (monthly and quarterly), ECL~\cite{zhou2021informer}, Exchange~\cite{lai2018modeling}, Hospital~\cite{godahewa2021monash}, Traffic~\cite{godahewa2021monash}, and Weather~\cite{zhou2021informer}. 
Traffic and ECL were originally collected at an hourly frequency, and Weather at a 10-minute frequency; we resample these datasets to daily frequency by taking the mean to improve computational efficiency and make the experiments more tractable. The datasets are described in Table~\ref{tab:data}.

\begin{table}[t]
  \caption{Summary of the datasets used in the experimental setup, including the number of time series, number of observations, forecast horizon, and frequency.}
  \label{tab:data}
  \centering
  \begin{tabular}{llrrr}
    \toprule
    Dataset & Frequency & \# time series & \# observations & Horizon ($h$) \\
    \midrule
    \multirow{2}{*}{Tourism} & Monthly   & 366    & 109,280 & 12 \\
    & Quarterly & 427    & 42,544   & 4  \\
    \midrule
    \multirow{2}{*}{M1} & Monthly   & 617    & 55,998  & 12 \\
                       & Quarterly & 203    & 9,944   & 4  \\
    \midrule
    \multirow{2}{*}{M3} & Monthly   & 1,428  & 167,562 & 12 \\
                       & Quarterly & 756    & 37,004  & 4  \\
    \midrule
    Exchange & Daily & 8    & 60,704  & 14 \\
    Weather & Daily & 21    & 7,707  & 14 \\
    Traffic & Daily & 862    & 630,984  & 14 \\
    ECL & Daily & 321    & 352,137  & 14 \\
    Hospital      & Monthly & 767 & 64,428 & 12 \\
    \midrule
    \textbf{Total} & & \textbf{5,776} & \textbf{1,538,292} & -- \\
    \bottomrule
  \end{tabular}%
\end{table}

\subsection{Methods}\label{sec:learning_algorithms}

For forecasting models we consider 10 deep learning architectures: \texttt{MLP}, \texttt{DLinear}~\cite{zeng2023transformers}, \texttt{NLinear}~\cite{zeng2023transformers}, \texttt{TiDE}~\cite{das2023long}, \texttt{NHITS}~\cite{challu2023nhits}, \texttt{NBEATS}~\cite{oreshkin2019n}, \texttt{KAN}~\cite{liu2024kan}, \texttt{DeepNPTS}~\cite{rangapuram2023deep}, \texttt{PatchTST}~\cite{nie2022time}, and \texttt{TFT}~\cite{lim2021temporal}. For each model, hyperparameters are tuned with 20 trials of random search within the nested CV procedure described above. Complete model-specific hyperparameter search spaces and fixed random seeds are provided in the supplementary code repository (\url{https://github.com/vcerqueira/experiments-tscv}).

We remark that our goal is to compare CV techniques, not to benchmark neural architectures against one another. We therefore prioritize well-established, computationally efficient architectures that allow large-scale experimentation. The architecture pool should be viewed as a representative testbed for evaluating CV methodologies rather than a ranking of the most accurate forecasters.

We benchmark all series-wise CV methods listed in Section~\ref{sec:swcv_listofstrats}, along with the standard time-wise CV approach (\texttt{Time-CV}) which is typically used in time series forecasting (e.g.~\cite{oreshkin2019n,lim2021temporal,nie2022time,zeng2023transformers,das2023long}). The number of folds for \texttt{K-Holdout}, \texttt{K-MC-CV}, \texttt{K-Bootstrap}, and \texttt{Kf-CV} is set to 5. In \texttt{Holdout} and \texttt{K-Holdout}, the percentage of the training set is 70\% of the time series. For each fold in \texttt{K-MC-CV}, the training and testing sets are composed of 50\% and 20\% of the time series, while the remaining 30\% are unused. 

\section{Results}

This section presents the results of the experiments, starting with the scores for performance estimation and model selection (Section~\ref{sec:results1}) and then the scores for forecasting accuracy (Section~\ref{sec:results2}).

\subsection{Performance Estimation and Model Selection}\label{sec:results1}

Table~\ref{tab:estimation} shows the results on performance estimation and model selection.
\texttt{Time-CV} shows the best performance estimation error (MAPEE), followed by \texttt{Bootstrap} and \texttt{Kf-CV}. 
\texttt{Holdout} and \texttt{K-MC-CV} show the largest MAPEE.
Overoptimism ranks the methods differently. \texttt{K-Bootstrap} and \texttt{Kf-CV} are the least overoptimistic approaches ($\sim$0.26), while \texttt{K-Holdout} and \texttt{K-MC-CV} also stay near that score. \texttt{Time-CV} (0.324) under-estimates the loss more often than each of these series-wise procedures. 

\begin{table}[t]
\caption{Results of performance estimation (MAPEE), overoptimism rate, selection accuracy, and regret for each CV approach. Numbers in bold (underlined) denote the best (second-best) approach.}
\label{tab:estimation}
\resizebox{\textwidth}{!}{%
\begin{tabular}{llllllll}
\multicolumn{7}{c}{Series-wise} & Time-wise \\
\cmidrule(r){2-7} \cmidrule(l){8-8}
\toprule
Method & \texttt{Bootstrap} & \texttt{Holdout} & \texttt{Kf-CV} & \texttt{K-MC-CV} & \texttt{K-Bootstrap} & \texttt{K-Holdout} & \texttt{Time-CV} \\
\midrule
MAPEE & \underline{0.120} & 0.141 & 0.124 & 0.140 & 0.126 & 0.132 & \textbf{0.109} \\
Overoptimism & 0.315 & 0.373 & \underline{0.258} & 0.267 & \textbf{0.255} & 0.261 & 0.324 \\
Accuracy & 0.242 & \textbf{0.364} & 0.242 & \underline{0.303} & 0.152 & 0.212 & 0.273 \\
Regret & 0.023 & \underline{0.019} & 0.044 & \textbf{0.016} & \underline{0.019} & 0.020 & 0.034 \\
\bottomrule
\end{tabular}%
}
\end{table}

Regarding model selection, the accuracy of the estimators ranges between 15.2\% and 36.4\%, which is better than random selection (1 over 10 alternatives, which equals to 10\% accuracy on average). \texttt{Holdout} shows the highest accuracy, followed by \texttt{K-MC-CV}.
Concerning the regret of each CV procedure, \texttt{K-MC-CV} also shows the best score, followed by \texttt{Holdout} and \texttt{K-Bootstrap}.
These results suggest that time-wise partitioning gives the smallest absolute error in the performance estimate, while series-wise procedures under-estimate that error less often. Series-wise CV also provides competitive alternatives for model selection based on the selection accuracy and regret scores.

\subsection{Downstream Forecasting Accuracy}\label{sec:results2}

\begin{table}[tb]
\caption{MASE score of each CV approach on each dataset, and respective average rank, average MASE, and top 2 counts (averaged across seeds).}
\label{tab:forecasting-seeds}
\resizebox{\textwidth}{!}{%
\begin{tabular}{llllllll}
\multicolumn{7}{c}{Series-wise} & Time-wise \\
\cmidrule(r){2-7} \cmidrule(l){8-8}
\toprule
Methods & \texttt{Bootstrap} & \texttt{Holdout} & \texttt{Kf-CV} & \texttt{K-MC-CV} & \texttt{K-Bootstrap} & \texttt{K-Holdout} & \texttt{Time-CV} \\
\midrule
ECL & 0.596 & 0.592 & \underline{0.591} & 0.595 & 0.600 & 0.604 & \textbf{0.577} \\
Exchange & 0.151 & 0.153 & \underline{0.147} & \textbf{0.145} & \textbf{0.145} & \underline{0.147} & 0.155 \\
Traffic & 0.702 & \textbf{0.671} & 0.688 & 0.691 & 0.691 & \underline{0.677} & 0.692 \\
Weather & 0.976 & \textbf{0.920} & 0.939 & \underline{0.928} & 0.940 & 0.934 & 1.115 \\
Hospital & 0.811 & 0.814 & \underline{0.797} & \textbf{0.796} & 0.804 & 0.804 & 0.804 \\
M1-M & 0.904 & \textbf{0.901} & 0.909 & 0.912 & \underline{0.903} & \underline{0.903} & 0.920 \\
M1-Q & 1.124 & 1.138 & \underline{1.120} & 1.131 & 1.121 & \textbf{1.114} & 1.124 \\
M3-M & 0.813 & 0.797 & \underline{0.755} & 0.799 & 0.797 & 0.786 & \textbf{0.737} \\
M3-Q & 0.832 & \underline{0.827} & 0.891 & \textbf{0.819} & 0.848 & 0.865 & 0.874 \\
Tourism-M & \underline{1.320} & \underline{1.320} & \textbf{1.312} & 1.333 & 1.333 & 1.321 & 1.323 \\
Tourism-Q & \textbf{1.275} & 1.304 & 1.494 & \underline{1.278} & 1.307 & 1.290 & 1.387 \\
\midrule
Avg. Rank & 4.303 & \textbf{3.545} & \underline{3.561} & 3.712 & 4.030 & 3.970 & 4.879 \\
Avg & 0.864 & \underline{0.858} & 0.877 & \textbf{0.857} & 0.863 & 0.859 & 0.883 \\
Top 2 Count & 3.667 & \underline{5.000} & \textbf{5.667} & 4.333 & 3.333 & \underline{5.000} & 2.000 \\
\bottomrule
\end{tabular}%
}
\end{table}

Table~\ref{tab:forecasting-seeds} shows the downstream forecasting performance with each CV procedure as part of the forecasting pipeline (cf. Section~\ref{sec:performance_estimation_evaluation}). In this scenario, time-wise CV shows the worst average rank, being outperformed by all series-wise CV variants. \texttt{Holdout} shows the best average rank of 3.545 and the second best average MASE (0.858), and attains the top 2 score for 5 out of the 11 datasets. These results support the hypothesis that series-wise CV can improve downstream forecasting relative to time-wise CV. 
It is worth noting that \texttt{Holdout} is an approach that conducts a single holdout split of the in-set to select the model, rather than a $K$-fold procedure. 
As mentioned before, the scores presented above are the average of the 3 seeds. For further details, the per-seed results are available in the appendix (Tables~\ref{tab:forecasting-seed-123}--\ref{tab:estimation-seed-789}). In those, \texttt{Holdout} shows the best forecasting performance for 2 out of the 3 seeds, with \texttt{K-Holdout} showing the best performance for the other seed. Across all seeds, time-wise CV systematically underperforms the series-wise CV variants in downstream forecasting accuracy.

\subsection{Results with a Complete In-set}

The results so far employ an evaluation protocol that holds out a random fraction of series from the in-set so that they appear only in the out-set. This protocol simulates the emergence of new series not seen during training, a common scenario in global forecasting. In this section, we also present the results where the in-set contains all time series.
The results are shown in Table~\ref{tab:estimation-tw} and Table~\ref{tab:forecasting-tw}. 

\begin{table}[t]
\caption{Results of performance estimation (MAPEE), overoptimism rate, selection accuracy, and regret for each CV approach with a complete in-set.}
\label{tab:estimation-tw}
\resizebox{\textwidth}{!}{%
\begin{tabular}{llllllll}
\multicolumn{7}{c}{Series-wise} & Time-wise \\
\cmidrule(r){2-7} \cmidrule(l){8-8}
\toprule
Method & \texttt{Bootstrap} & \texttt{Holdout} & \texttt{Kf-CV} & \texttt{K-MC-CV} & \texttt{K-Bootstrap} & \texttt{K-Holdout} & \texttt{Time-CV} \\
\midrule
MAPEE & 0.119 & 0.109 & 0.108 & 0.127 & 14.321 & \underline{0.105} & \textbf{0.092} \\
Overoptimism & 0.306 & 0.424 & 0.255 & \underline{0.252} & \textbf{0.242} & 0.261 & 0.306 \\
Accuracy & \textbf{0.455} & 0.273 & 0.273 & 0.273 & 0.273 & \underline{0.364} & 0.333 \\
Regret & 0.025 & 0.024 & 0.022 & 0.020 & \underline{0.019} & 0.020 & \textbf{0.012} \\
\bottomrule
\end{tabular}%
}
\end{table}

When all series appear in the in-set, the forecasting comparison changes. \texttt{Time-CV} attains the best MAPEE and regret and the best forecasting accuracy. On overoptimism, time-wise CV is outperformed by several series-wise approaches, as in Table~\ref{tab:estimation}. Selection accuracy is highest for \texttt{Bootstrap}, with time-wise CV being third best. 

Overall, the forecasting advantage of series-wise CV disappears once no new series emerge at test time, while its tendency for overoptimism remains.
This suggests that the choice of CV procedure should match the expected population dynamics. Series-wise CV is preferable when the population of series is expected to evolve and new series appear at test time. Time-wise CV remains the better choice for downstream accuracy when the population is static. When under-estimating the error is costly, series-wise procedures are effective approaches in both regimes.

\begin{table}[tb]
\caption{MASE score of each CV approach on each dataset, and respective average rank, average MASE, and top 2 counts with a complete in-set.}
\label{tab:forecasting-tw}
\resizebox{\textwidth}{!}{%
\begin{tabular}{llllllll}
\multicolumn{7}{c}{Series-wise} & Time-wise \\
\cmidrule(r){2-7} \cmidrule(l){8-8}
\toprule
Methods & \texttt{Bootstrap} & \texttt{Holdout} & \texttt{Kf-CV} & \texttt{K-MC-CV} & \texttt{K-Bootstrap} & \texttt{K-Holdout} & \texttt{Time-CV} \\
\midrule
ECL & 0.592 & \textbf{0.583} & \underline{0.587} & 0.598 & 0.594 & 0.596 & 0.594 \\
Exchange & 0.147 & \textbf{0.142} & 0.148 & 0.148 & \underline{0.143} & 0.144 & 0.156 \\
Traffic & 0.686 & 0.701 & 0.686 & \underline{0.684} & 0.687 & \textbf{0.683} & 0.686 \\
Weather & \textbf{0.922} & 0.936 & 0.930 & \underline{0.926} & \underline{0.926} & 0.933 & 0.941 \\
Hospital & 0.813 & \textbf{0.797} & 0.801 & \textbf{0.797} & 0.808 & \textbf{0.797} & \underline{0.800} \\
M1-M & 0.907 & \underline{0.906} & 0.916 & 0.910 & 0.919 & 0.921 & \textbf{0.904} \\
M1-Q & 1.276 & 1.209 & \textbf{1.124} & 1.196 & 1.186 & 1.175 & \underline{1.127} \\
M3-M & 0.779 & 0.763 & 0.762 & 0.778 & 0.780 & \textbf{0.750} & \underline{0.761} \\
M3-Q & 0.860 & 0.842 & 0.845 & 0.838 & \underline{0.832} & \textbf{0.821} & \textbf{0.821} \\
Tourism-M & \underline{1.316} & 1.334 & 1.335 & \underline{1.316} & 1.327 & 1.321 & \textbf{1.303} \\
Tourism-Q & 1.288 & 1.290 & 1.336 & 1.256 & \underline{1.255} & 1.340 & \textbf{1.244} \\
\midrule
Avg. Rank & 4.076 & 4.091 & 4.364 & \underline{3.909} & 4.167 & 3.939 & \textbf{3.455} \\
Avg & 0.871 & 0.864 & 0.861 & \underline{0.859} & 0.860 & 0.862 & \textbf{0.849} \\
Top 2 Count & \textbf{4.667} & 3.667 & 2.333 & 3.667 & 3.667 & 3.667 & \underline{4.333} \\
\bottomrule
\end{tabular}%
}
\end{table}

\section{Discussion}

\subsection{Key findings} 

This paper formalizes series-wise CV for global time series forecasting, partitioning data across both series and time, and empirically compares several instantiations against the prevailing time-wise CV baseline. Our experiments span 11 datasets and 10 neural architectures, evaluating CV procedures along different dimensions: performance estimation accuracy, model selection quality, and downstream forecasting performance. The key findings
are:

\begin{itemize}

    \item \textbf{Performance estimation.} Time-wise CV yields the lowest estimation error (Table~\ref{tab:estimation}), indicating that temporal partitioning remains the most reliable strategy for approximating out-of-sample loss. However, time-wise CV also tends to produce overoptimistic estimates more often than several series-wise CV procedures.

    \item \textbf{Model selection and regret.} Series-wise methods, notably \texttt{Holdout} and \texttt{K-MC-CV}, achieve higher selection accuracy and lower regret than time-wise CV, showing that optimizing for generalization across series can improve the choice of model.

    \item \textbf{Downstream forecasting.} When CV is embedded in the full forecasting pipeline, almost all series-wise CV variants outperform time-wise CV on average rank and average MASE (Table~\ref{tab:forecasting-seeds}). This outcome supports our central hypothesis: validating on unseen series steers selection toward configurations that generalize across both time and the space of series. 

    \item \textbf{Dynamics of the population of time series.} Series-wise CV delivers better downstream accuracy when new series appear at test time. When only previously seen series remain in the population, time-wise CV is the stronger choice for downstream forecasting accuracy. 
\end{itemize}

One important result from our experiments is that \emph{accurate performance estimation does not imply accurate model selection}, corroborating previous works in this topic~\cite{cerqueira2023model}. Time-wise CV produces estimates closer to the true out-of-sample error in absolute value, and series-wise procedures under-estimate that error less often. Series-wise CV also more often identifies the best-performing model. This outcome can be understood by the fact that time-wise CV rewards models that extrapolate well on familiar dynamics, while series-wise CV rewards models that transfer to unseen patterns. When the goal is to deploy a model that performs robustly across an evolving population of series, which is the typical use case for global forecasters, including time series foundation models, the latter objective appears more aligned with downstream success.

For practitioners building global forecasters, our results suggest the adoption of series-wise CV, particularly \texttt{Holdout}, \texttt{K-Holdout} or \texttt{K-MC-CV}, when the primary goal is downstream accuracy, and the use of time-wise CV when the magnitude of the error estimate matters most. When under-estimating the error is an important concern, series-wise procedures are also an effective choice. 

\subsection{Limitations and future work}

Series-wise CV methods involving $K$ folds incur roughly $K$-fold additional training time. However, this overhead is confined to offline model development and does not affect inference latency. Moreover, folds can be executed in parallel, and our results show that even single-split series-wise variants ($K{=}1$) like \texttt{Holdout} outperform time-wise CV on downstream accuracy, offering a low-cost entry point for practitioners with limited compute budgets.

It is also worth noting that we use a simplified time-wise CV in the inner loop of nested series-wise CV, mirroring common practice where the inner procedure is simpler than the outer loop~\cite{arlot2010survey}. On the other hand, this choice also isolates the effect of the outer CV strategy, which supports a fair comparison.

\section{Conclusions}\label{sec:conclusions}

CV for time series forecasting has traditionally partitioned data across time alone. The shift toward global models, trained jointly on many series, opens a complementary axis: partitioning across series. In this paper, we formalized \emph{series-wise} CV for global forecasting models, presented several instantiations, and benchmarked them against the prevailing time-wise baseline on 11 datasets and 10 neural architectures, along five dimensions spanning performance estimation, model selection, and downstream forecasting.

Series-wise CV is most valuable in the setting that increasingly defines global forecasting. When new series emerge at test time, it selects models that forecast more accurately downstream as it rewards transfer to unseen series rather than only temporal extrapolation on familiar ones. Global forecasters are often typically deployed in exactly this regime, on populations in which new series continue to appear. Time-wise CV remains the more reliable way to approximate the magnitude of out-of-sample error, and it is also preferable for downstream accuracy when the population of series is expected to remain static. Series-wise procedures under-estimate that error less often, in both the evolving and the static population.

The experimental results also point to several directions for future work. Our evaluation emulates an evolving population through a series-wise out-set in which a fraction of series are unseen during training; studying richer forms of distribution shift, such as gradual drift within series or varying fractions of emerging series, would further delineate when series-wise CV helps. We release our benchmark code to support these extensions and to facilitate the adoption of series-wise CV.

\section*{Acknowledgements}

\small{
This work is funded by national funds through FCT – Foundation for Science and Technology, I.P., within the scope of the research unit UID/00326 - Centre for Informatics and Systems of the University of Coimbra, \url{https://doi.org/10.54499/UID/00326/2025}.
This work was partially funded by project AISym4Med (101095387) supported by Horizon Europe Cluster 1: Health, ConnectedHealth (n.o 46858), supported by Competitiveness and Internationalisation Operational Programme (POCI) and Lisbon Regional Operational Programme (LISBOA 2020), under the PORTUGAL 2020 Partnership Agreement, through the European Regional Development Fund (ERDF). AISym4Med is funded by the European Union under the Horizon Europe Framework Programme Grant Agreement Nº: 101095387. Views and opinions expressed are however those of the author(s) only and do not necessarily reflect those of the European Union or European Commission. Neither the European Union nor the European Health and Digital Executive Agency can be held responsible for them.}

\vskip 0.2in
\bibliography{references}

\clearpage
\appendix

\section{Results by Seed}
\label{app:seed}

The main body of the paper shows the results averaged across three seeds: \textit{123}, \textit{456}, \textit{789}. For transparency, Tables~\ref{tab:forecasting-seed-123}--~\ref{tab:estimation-seed-789} show the per-seed scores across all five dimensions. In general, the key conclusions remain. Time-wise CV is outperformed by almost all series-wise CV variants on downstream forecasting performance. On MAPEE it is best or second-best in each seed. The best approach for forecasting accuracy is not the same in every run: \texttt{Holdout} leads in two of the three seeds, and \texttt{K-Holdout} leads in the remaining one.

\begin{table}[tb]
\caption{MASE score of each CV approach on each dataset, and respective average rank, average MASE, and top 2 counts (seed 123).}
\label{tab:forecasting-seed-123}
\resizebox{\textwidth}{!}{%
\begin{tabular}{llllllll}
\multicolumn{7}{c}{Series-wise} & Time-wise \\
\cmidrule(r){2-7} \cmidrule(l){8-8}
\toprule
Methods & \texttt{Bootstrap} & \texttt{Holdout} & \texttt{Kf-CV} & \texttt{K-MC-CV} & \texttt{K-Bootstrap} & \texttt{K-Holdout} & \texttt{Time-CV} \\
\midrule
ECL & 0.586 & 0.595 & \underline{0.585} & 0.586 & 0.586 & 0.606 & \textbf{0.574} \\
Exchange & 0.152 & 0.168 & \textbf{0.141} & 0.146 & \underline{0.143} & 0.145 & 0.159 \\
Traffic & 0.700 & \textbf{0.675} & \underline{0.690} & 0.698 & \underline{0.690} & \underline{0.690} & 0.695 \\
Weather & 1.063 & \textbf{0.921} & \underline{0.939} & \underline{0.939} & \underline{0.939} & \underline{0.939} & 1.148 \\
Hospital & 0.806 & 0.807 & \underline{0.800} & \textbf{0.796} & 0.806 & 0.821 & 0.808 \\
M1-M & \underline{0.910} & \textbf{0.901} & \textbf{0.901} & \underline{0.910} & \textbf{0.901} & \textbf{0.901} & 0.932 \\
M1-Q & 1.125 & \textbf{1.115} & 1.123 & 1.154 & 1.125 & \textbf{1.115} & \underline{1.120} \\
M3-M & 0.843 & 0.779 & \underline{0.751} & 0.779 & 0.773 & 0.782 & \textbf{0.737} \\
M3-Q & 0.858 & 0.830 & 0.906 & \textbf{0.817} & 0.858 & \underline{0.823} & 0.896 \\
Tourism-M & \textbf{1.296} & \textbf{1.296} & \textbf{1.296} & 1.321 & 1.359 & \underline{1.320} & 1.323 \\
Tourism-Q & 1.309 & \textbf{1.239} & 1.914 & \underline{1.275} & 1.290 & 1.309 & 1.434 \\
\midrule
Avg. Rank & 5.000 & \textbf{3.136} & \underline{3.273} & 3.909 & 3.773 & 4.000 & 4.909 \\
Avg & 0.877 & \textbf{0.848} & 0.913 & \underline{0.856} & 0.861 & 0.859 & 0.893 \\
Top 2 Count & 1 & \underline{6} & \textbf{8} & 4 & 4 & 5 & 2 \\
\bottomrule
\end{tabular}%
}
\end{table}

\begin{table}[tb]
\caption{MASE score of each CV approach on each dataset, and respective average rank, average MASE, and top 2 counts (seed 456).}
\label{tab:forecasting-seed-456}
\resizebox{\textwidth}{!}{%
\begin{tabular}{llllllll}
\multicolumn{7}{c}{Series-wise} & Time-wise \\
\cmidrule(r){2-7} \cmidrule(l){8-8}
\toprule
Methods & \texttt{Bootstrap} & \texttt{Holdout} & \texttt{Kf-CV} & \texttt{K-MC-CV} & \texttt{K-Bootstrap} & \texttt{K-Holdout} & \texttt{Time-CV} \\
\midrule
ECL & 0.598 & \underline{0.595} & 0.602 & \textbf{0.583} & 0.598 & 0.598 & \textbf{0.583} \\
Exchange & 0.151 & \underline{0.142} & 0.145 & 0.149 & \textbf{0.141} & \textbf{0.141} & 0.147 \\
Traffic & \underline{0.686} & \textbf{0.684} & \textbf{0.684} & \textbf{0.684} & \textbf{0.684} & \underline{0.686} & 0.687 \\
Weather & 0.945 & \textbf{0.898} & 0.940 & \underline{0.920} & 0.942 & \underline{0.920} & 1.050 \\
Hospital & 0.840 & 0.813 & \textbf{0.795} & 0.797 & 0.801 & \underline{0.796} & 0.803 \\
M1-M & \underline{0.902} & \underline{0.902} & 0.925 & 0.925 & 0.907 & 0.907 & \textbf{0.895} \\
M1-Q & \underline{1.123} & 1.184 & \underline{1.123} & \textbf{1.113} & \underline{1.123} & \textbf{1.113} & 1.131 \\
M3-M & 0.846 & 0.839 & \textbf{0.736} & 0.839 & 0.839 & \textbf{0.736} & \underline{0.737} \\
M3-Q & \textbf{0.816} & 0.827 & 0.908 & \underline{0.823} & 0.827 & 0.866 & 0.831 \\
Tourism-M & \textbf{1.301} & 1.326 & \underline{1.321} & 1.357 & \underline{1.321} & \underline{1.321} & 1.323 \\
Tourism-Q & \textbf{1.248} & 1.404 & 1.292 & \underline{1.284} & 1.340 & \textbf{1.248} & 1.293 \\
\midrule
Avg. Rank & 4.364 & 4.227 & 3.955 & \underline{3.636} & 4.045 & \textbf{3.136} & 4.636 \\
Avg & \underline{0.860} & 0.874 & 0.861 & 0.861 & 0.866 & \textbf{0.848} & 0.862 \\
Top 2 Count & 4 & 3 & 4 & \underline{5} & 3 & \textbf{7} & 2 \\
\bottomrule
\end{tabular}%
}
\end{table}

\begin{table}[tb]
\caption{MASE score of each CV approach on each dataset, and respective average rank, average MASE, and top 2 counts (seed 789).}
\label{tab:forecasting-seed-789}
\resizebox{\textwidth}{!}{%
\begin{tabular}{llllllll}
\multicolumn{7}{c}{Series-wise} & Time-wise \\
\cmidrule(r){2-7} \cmidrule(l){8-8}
\toprule
Methods & \texttt{Bootstrap} & \texttt{Holdout} & \texttt{Kf-CV} & \texttt{K-MC-CV} & \texttt{K-Bootstrap} & \texttt{K-Holdout} & \texttt{Time-CV} \\
\midrule
ECL & 0.604 & 0.586 & \underline{0.585} & 0.615 & 0.615 & 0.606 & \textbf{0.574} \\
Exchange & 0.152 & \underline{0.148} & 0.154 & \textbf{0.142} & 0.152 & 0.154 & 0.159 \\
Traffic & 0.721 & \textbf{0.654} & \underline{0.690} & \underline{0.690} & 0.700 & \textbf{0.654} & 0.695 \\
Weather & \textbf{0.920} & 0.939 & 0.939 & \underline{0.926} & 0.939 & 0.943 & 1.148 \\
Hospital & \textbf{0.787} & 0.821 & \underline{0.796} & \underline{0.796} & 0.803 & \underline{0.796} & 0.802 \\
M1-M & \textbf{0.901} & \textbf{0.901} & \textbf{0.901} & \textbf{0.901} & \textbf{0.901} & \underline{0.902} & 0.932 \\
M1-Q & 1.125 & \textbf{1.115} & \textbf{1.115} & 1.125 & \textbf{1.115} & \textbf{1.115} & \underline{1.120} \\
M3-M & \underline{0.751} & 0.773 & 0.779 & 0.779 & 0.779 & 0.841 & \textbf{0.737} \\
M3-Q & \underline{0.823} & \underline{0.823} & 0.858 & \textbf{0.817} & 0.858 & 0.906 & 0.896 \\
Tourism-M & 1.365 & 1.339 & \textbf{1.320} & \underline{1.321} & \textbf{1.320} & \underline{1.321} & 1.323 \\
Tourism-Q & \textbf{1.268} & \textbf{1.268} & \underline{1.275} & \underline{1.275} & 1.290 & 1.313 & 1.434 \\
\midrule
Avg. Rank & 3.545 & \textbf{3.273} & \underline{3.455} & 3.591 & 4.273 & 4.773 & 5.091 \\
Avg & 0.856 & \textbf{0.851} & 0.856 & \underline{0.853} & 0.861 & 0.868 & 0.893 \\
Top 2 Count & \textbf{6} & \textbf{6} & \underline{5} & 4 & 3 & 3 & 2 \\
\bottomrule
\end{tabular}%
}
\end{table}

\begin{table}[t]
\caption{Results of performance estimation (MAPEE), overoptimism rate, selection accuracy, and regret for each CV approach (seed 123).}
\label{tab:estimation-seed-123}
\resizebox{\textwidth}{!}{%
\begin{tabular}{llllllll}
\multicolumn{7}{c}{Series-wise} & Time-wise \\
\cmidrule(r){2-7} \cmidrule(l){8-8}
\toprule
Method & \texttt{Bootstrap} & \texttt{Holdout} & \texttt{Kf-CV} & \texttt{K-MC-CV} & \texttt{K-Bootstrap} & \texttt{K-Holdout} & \texttt{Time-CV} \\
\midrule
MAPEE & \textbf{0.105} & 0.166 & 0.133 & 0.135 & 0.136 & 0.129 & \underline{0.112} \\
Overoptimism & 0.282 & 0.336 & \underline{0.245} & 0.336 & \textbf{0.227} & 0.255 & 0.327 \\
Accuracy & 0.091 & \textbf{0.455} & \underline{0.364} & 0.091 & 0.182 & 0.273 & 0.182 \\
Regret & 0.031 & \textbf{0.008} & 0.073 & 0.021 & 0.022 & \underline{0.018} & 0.043 \\
\bottomrule
\end{tabular}%
}
\end{table}

\begin{table}[t]
\caption{Results of performance estimation (MAPEE), overoptimism rate, selection accuracy, and regret for each CV approach (seed 456). }
\label{tab:estimation-seed-456}
\resizebox{\textwidth}{!}{%
\begin{tabular}{llllllll}
\multicolumn{7}{c}{Series-wise} & Time-wise \\
\cmidrule(r){2-7} \cmidrule(l){8-8}
\toprule
Method & \texttt{Bootstrap} & \texttt{Holdout} & \texttt{Kf-CV} & \texttt{K-MC-CV} & \texttt{K-Bootstrap} & \texttt{K-Holdout} & \texttt{Time-CV} \\
\midrule
MAPEE & 0.134 & \underline{0.118} & 0.120 & 0.157 & 0.121 & 0.133 & \textbf{0.103} \\
Overoptimism & 0.336 & 0.4 & 0.291 & \textbf{0.236} & \underline{0.264} & \underline{0.264} & 0.309 \\
Accuracy & \underline{0.364} & 0.273 & 0.182 & \underline{0.364} & 0.182 & 0.273 & \textbf{0.455} \\
Regret & 0.019 & 0.035 & 0.028 & \textbf{0.014} & \underline{0.015} & 0.017 & \textbf{0.014} \\
\bottomrule
\end{tabular}%
}
\end{table}

\begin{table}[t]
\caption{Results of performance estimation (MAPEE), overoptimism rate, selection accuracy, and regret for each CV approach (seed 789). }
\label{tab:estimation-seed-789}
\resizebox{\textwidth}{!}{%
\begin{tabular}{llllllll}
\multicolumn{7}{c}{Series-wise} & Time-wise \\
\cmidrule(r){2-7} \cmidrule(l){8-8}
\toprule
Method & \texttt{Bootstrap} & \texttt{Holdout} & \texttt{Kf-CV} & \texttt{K-MC-CV} & \texttt{K-Bootstrap} & \texttt{K-Holdout} & \texttt{Time-CV} \\
\midrule
MAPEE & 0.122 & 0.140 & \underline{0.119} & 0.127 & 0.122 & 0.132 & \textbf{0.114} \\
Overoptimism & 0.327 & 0.382 & \underline{0.236} & \textbf{0.227} & 0.273 & 0.264 & 0.336 \\
Accuracy & 0.273 & \underline{0.364} & 0.182 & \textbf{0.455} & 0.091 & 0.091 & 0.182 \\
Regret & 0.019 & \underline{0.016} & 0.032 & \textbf{0.012} & 0.020 & 0.026 & 0.043 \\
\bottomrule
\end{tabular}%
}
\end{table}

\end{document}